\def\year{2022}\relax
\documentclass[letterpaper]{article} % DO NOT CHANGE THIS
\usepackage{aaai22}  % DO NOT CHANGE THIS
\usepackage{times}  % DO NOT CHANGE THIS
\usepackage{helvet} % DO NOT CHANGE THIS
\usepackage{courier}  % DO NOT CHANGE THIS
\usepackage[hyphens]{url}  % DO NOT CHANGE THIS
\usepackage{graphicx} % DO NOT CHANGE THIS
\usepackage{natbib}  % DO NOT CHANGE THIS AND DO NOT ADD ANY OPTIONS TO IT
\usepackage{caption} % DO NOT CHANGE THIS AND DO NOT ADD ANY OPTIONS TO IT
\usepackage{booktabs}
\usepackage{multirow}

\usepackage{amsmath, amssymb}
\usepackage[pagebackref=true,breaklinks=true,letterpaper=true,colorlinks,bookmarks=false]{hyperref}
\makeatletter
\newcommand*\mysize{%
  \@setfontsize\mysize{8.0}{9.0}%
}
\makeatother

\newcommand{\similarity}{\operatornamewithlimits{sim}}

\title{Semantic Feature Extraction for Generalized Zero-shot Learning}
\author{
    Junhan Kim, Kyuhong Shim, Byonghyo Shim \\
}
\affiliations{
    Department of Electrical and Computer Engineering, Seoul National University, Seoul, Korea \\
    \{junhankim, khshim, bshim\}@islab.snu.ac.kr

}
\begin{document}

\maketitle

\begin{abstract}
    Generalized zero-shot learning (GZSL) is a technique to train a deep learning model to identify unseen classes using the attribute. 
    In this paper, we put forth a new GZSL technique that improves the GZSL classification performance greatly.
    Key idea of the proposed approach, henceforth referred to as semantic feature extraction-based GZSL (SE-GZSL), is to use the semantic feature containing only attribute-related information in learning the relationship between the image and the attribute.
    In doing so, we can remove the interference, if any, caused by the attribute-irrelevant information contained in the image feature.
    To train a network extracting the semantic feature, we present two novel loss functions, 1) mutual information-based loss to capture all the attribute-related information in the image feature and 2) similarity-based loss to remove unwanted attribute-irrelevant information.
    From extensive experiments using various datasets, we show that the proposed SE-GZSL technique outperforms conventional GZSL approaches by a large margin.
\end{abstract}

\section{Introduction}

Image classification is a long-standing yet important task with a wide range of applications such as autonomous driving, industrial automation, medical diagnosis, and biometric identification~\cite{autonomous_driving, industrial_automation, medical_diagnosis, biometric_identification}.
In solving the task, supervised learning (SL) techniques have been popularly used for its superiority~\cite{VGG, ResNet}.
Well-known drawback of SL is that a large number of training data are required for each and every class to be identified.
Unfortunately, in many practical scenarios, it is difficult to collect training data for certain classes (e.g., endangered species and newly observed species such as variants of COVID-19).
% When there are \textit{unseen} classes where training data is not available, SL-based models are biased towards the \textit{seen} classes observed in the training phase so that they cannot identify unseen classes.
When there are \textit{unseen} classes where training data is unavailable, SL-based models are biased towards the \textit{seen} classes, impeding the identification of the unseen classes.

\begin{figure}[!t] 
    \centering
    \centerline{\includegraphics[width=1\linewidth]{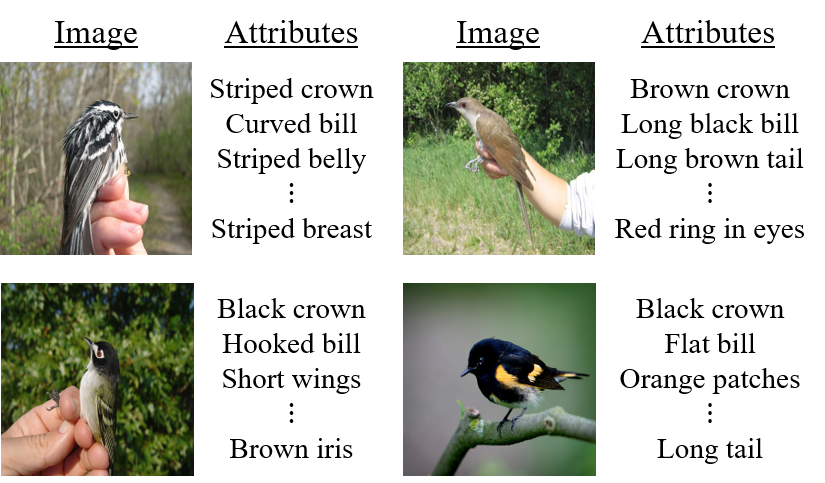}}
    \caption{Images and attributes for different bird species sampled from the CUB dataset~\cite{CUB}.}
    \label{fig:CUB}
\end{figure}

Recently, to overcome this drawback, a technique to train a classifier using manually annotated attributes (e.g., color, size, and shape; see Fig.~\ref{fig:CUB}) has been proposed~\cite{zsl_proposal, gzsl_intro}.
Key idea of this technique, dubbed as generalized zero-shot learning (GZSL), is to learn the relationship between the image and the attribute from seen classes and then use the trained model in the identification of unseen classes.
% In~\cite{DAPIAP}, for example, a network estimating the image attribute from the image feature has been used to classify images of unseen classes.
In~\cite{ALE}, for example, an approach to identify unseen classes by measuring the compatibility between the image feature and attribute has been proposed.
In~\cite{CVAE-GZSL}, a network synthesizing the image feature from the attribute has been employed to generate training data of unseen classes.
In extracting the image feature, a network trained using the classification task (e.g., ResNet~\cite{ResNet}) has been popularly used.
A potential drawback of this extraction method is that the image feature might contain attribute-irrelevant information (e.g., human fingers in Fig.~\ref{fig:CUB}), disturbing the process of learning the relationship between the image and the attribute~\cite{DLFZRL, RFF-GZSL, Disentangled-VAE}.

% A potential drawback of these approaches is that a pre-trained classification network (e.g., ResNet~\cite{ResNet}) has been used for the image feature extraction so that the image feature can contain the unwanted attribute-irrelevant information (e.g., human fingers in Fig.~\ref{fig:CUB})~\cite{DLFZRL}. 
% Since the attribute-irrelevant information can mess up the process to learn the relationship between the image feature and the attribute~\cite{DLFZRL, RFF-GZSL, Disentangled-VAE}, it is of importance to extract the image feature containing the attribute-related information exclusively.

\begin{figure*}[!t] 
    \centering
    \centerline{\includegraphics[width=15cm]{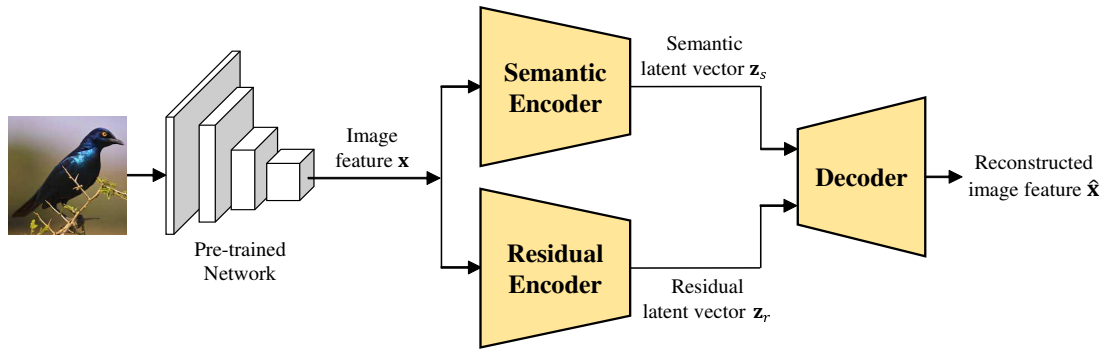}}
    \caption{Illustration of the image feature decomposition.}
    \label{fig:SD}
\end{figure*}

In this paper, we propose a new GZSL technique that removes the interference caused by the attribute-irrelevant information.
Key idea of the proposed approach is to extract the \textit{semantic feature}, feature containing the attribute-related information, from the image feature and then use it in learning the relationship between the image and the attribute.
In extracting the semantic feature, we use a modified autoencoder consisting of two encoders, viz., \textit{semantic} and \textit{residual} encoders (see Fig.~\ref{fig:SD}).
In a nutshell, the semantic encoder captures all the attribute-related information in the image feature and the residual encoder catches the attribute-irrelevant information.

% While the conventional approaches use image features containing attribute-irrelevant information for the classification, our approach, henceforth referred to as semantic feature extraction-based GZSL (SE-GZSL), exploits \textit{semantic features}, features containing only attribute-related information, to remove the interference caused by attribute-irrelevant information.

In the conventional autoencoder, only reconstruction loss (difference between the input and the reconstructed input) is used for the training.
In our approach, to encourage the semantic encoder to capture the attribute-related information only, we use two novel loss functions on top of the reconstruction loss.
% : 1) mutual information (MI)-based loss and 2) similarity-based loss.
% First, we employ the mutual information (MI)-based loss to enforce the semantic encoder to catch all the attribute-related information.
First, we employ the mutual information (MI)-based loss to maximize (minimize) MI between the semantic (residual) encoder output and the attribute.
Since MI is a metric to measure the level of dependency between two random variables, by exploiting the MI-based loss, we can encourage the semantic encoder to capture the attribute-related information and at the same time discourage the residual encoder to capture any attribute-related information.
As a result, all the attribute-related information can be solely captured by the semantic encoder.
Second, we use the similarity-based loss to enforce the semantic encoder not to catch any attribute-irrelevant information.
For example, when a bird image contains human fingers (see Fig.~\ref{fig:CUB}), we do not want features related to the finger to be included in the semantic encoder output.
To do so, we maximize the similarity between the semantic encoder outputs of images that are belonging to the same class (bird images in our example).
Since attribute-irrelevant features are contained only in a few image samples (e.g., human fingers are included in a few bird images), by maximizing the similarity between the semantic encoder outputs of the same class, we can remove attribute-irrelevant information from the semantic encoder output.

From extensive experiments using various benchmark datasets (AwA1, AwA2, CUB, and SUN), we demonstrate that the proposed approach outperforms the conventional GZSL techniques by a large margin.
For example, for the AwA2 and CUB datasets, our model achieves 2\% improvement in the GZSL classification accuracy over the state-of-the-art techniques.

\section{Related Work and Background}

\subsection{Conventional GZSL Approaches}

The main task in GZSL is to learn the relationship between the image and the attribute from seen classes and then use it in the identification of unseen classes.
Early GZSL works have focused on the training of a network measuring the compatibility score between the image feature and the attribute~\cite{ALE, DeViSE}.
Once the network is trained properly, images can be classified by identifying the attribute achieving the maximum compatibility score.
Recently, generative model-based GZSL approaches have been proposed~\cite{CVAE-GZSL, CLSWGAN}.
Key idea of these approaches is to generate synthetic image features of unseen classes from the attributes by employing a generative model~\cite{CVAE-GZSL, CLSWGAN}.
As a generative model, conditional variational autoencoder (CVAE)~\cite{vae} and conditional Wasserstein generative adversarial network (CWGAN)~\cite{WGAN} have been popularly used.
By exploiting the generated image features of unseen classes as training data, a classification network identifying unseen classes can be trained in a supervised manner.
% As a generative model to synthesize image features from attributes, variational autoencoder (VAE)~\cite{vae} and Wasserstein generative adversarial network (WGAN)~\cite{WGAN} have been popularly used.

Over the years, many efforts have been made to improve the performance of the generative model.
In~\cite{f-vaegan-d2, CADA-VAE, Zero-VAE-GAN}, an approach to combine multiple generative models (e.g., CVAE and CWGAN) has been proposed.
In~\cite{cycle-WGAN, DASCN}, an additional network estimating the image attribute from the image feature has been used to make sure that the synthetic image features satisfy the attribute of unseen classes.
In~\cite{CLSWGAN, LsrGAN, LisGAN}, an additional image classifier has been used in the generative model training to generate distinct image features for different classes.

Our approach is conceptually similar to the generative model-based approach in the sense that we generate synthetic image features of unseen classes using the generative model.
The key distinctive point of the proposed approach over the conventional approaches is that we use the features containing only attribute-related information in the classification to remove the interference, if any, caused by the attribute-irrelevant information. 

\subsection{MI for Deep Learning}

Mathematically, the MI $I(\mathbf{u}, \mathbf{v})$ between two random variables $\mathbf{u}$ and $\mathbf{v}$ is defined as
\begin{align}
    I(\mathbf{u}, \mathbf{v})
    &= \mathbb{E}_{p(\mathbf{u}, \mathbf{v})} \left [ \log \frac{p(\mathbf{u}, \mathbf{v})}{p(\mathbf{u}) p(\mathbf{v})} \right ] \nonumber \\
    &= \int_{\mathbf{u}} \int_{\mathbf{v}} p(\mathbf{u}, \mathbf{v}) \log \frac{p(\mathbf{u}, \mathbf{v})}{p(\mathbf{u}) p(\mathbf{v})} d \mathbf{u} d \mathbf{v},
    \label{eq:MI_definition}
\end{align}
where $p(\mathbf{u}, \mathbf{v})$ is the joint probability density function (PDF) of $\mathbf{u}$ and $\mathbf{v}$, and $p(\mathbf{u})$ and $p(\mathbf{v})$ are marginal PDFs of $\mathbf{u}$ and $\mathbf{v}$, respectively. 
% Recently, MI has been employed in deep learning to strengthen or weaken the independence between different parts of a neural network.
In practice, it is very difficult to compute the exact value of MI since the joint PDF $p(\mathbf{u}, \mathbf{v})$ is generally unknown and the integrals in~\eqref{eq:MI_definition} are often intractable.
To approximate MI, various MI estimators have been proposed~\cite{InfoNCE, CLUB}. 
Representative estimators include InfoNCE~\cite{InfoNCE} and contrastive log-ratio upper bound (CLUB)~\cite{CLUB}, defined as
\begin{align}
    I_{\text{InfoNCE}}(\mathbf{u}, \mathbf{v})
    % &= \mathbb{E}_{p(\mathbf{u}, \mathbf{v})} \hspace{-1mm}\left [ \log
    % \frac{\exp ( f(\mathbf{u}, \mathbf{v}) )}
    % {\mathbb{E}_{p(\mathbf{v}^{\prime})} \left [ \exp ( f(\mathbf{u}, \mathbf{v}^{\prime}) ) \right ]}  
    % \right ], \label{eq:MI_lower bound_InfoNCE} \\
    &= \mathbb{E}_{p(\mathbf{u}, \mathbf{v})} [ f(\mathbf{u}, \mathbf{v}) ] \hspace{-.5mm} \nonumber \\
    &~~~- \hspace{-.5mm} \mathbb{E}_{p(\mathbf{u})} \hspace{-1mm} \left [ \log \left (
    \mathbb{E}_{p(\mathbf{v})} [ \exp ( f(\mathbf{u}, \mathbf{v} ) ) ] 
    \right ) \right ], \label{eq:MI_lower bound_InfoNCE} \\
    I_{\text{CLUB}}(\mathbf{u}, \mathbf{v})
    &= \mathbb{E}_{p(\mathbf{u}, \mathbf{v})} \hspace{-1mm} \left [ \log p(\mathbf{v} | \mathbf{u} ) \right ] \hspace{-.5mm} - 
    \hspace{-.5mm} \mathbb{E}_{p(\mathbf{u})p(\mathbf{v})} \hspace{-1mm} \left [ \log p(\mathbf{v} | \mathbf{u} ) \right ],
    \label{eq:MI_upper bound_CLUB}
\end{align}
where $f$ is a pre-defined score function measuring the compatibility between $\mathbf{u}$ and $\mathbf{v}$, and $p(\mathbf{v} | \mathbf{u})$ is the conditional PDF of $\mathbf{v}$ given $\mathbf{u}$, which is often approximated by a neural network.

The relationship between MI, InfoNCE, and CLUB is given by
\begin{align}
    I_{\text{InfoNCE}}(\mathbf{u}, \mathbf{v})
    \le I(\mathbf{u}, \mathbf{v})
    \le I_{\text{CLUB}}(\mathbf{u}, \mathbf{v}).
    \label{eq:MI inequality}
\end{align}
Recently, InfoNCE and CLUB have been used to strengthen or weaken the independence between different parts of the neural network.
% Recently, these MI estimators have been employed to train the neural network.
For example, when one tries to enforce the independence between $\mathbf{u}$ and $\mathbf{v}$, that is, to reduce $I(\mathbf{u}, \mathbf{v})$, an approach to minimize the upper bound $I_{\text{CLUB}}(\mathbf{u}, \mathbf{v})$ of MI can be used~\cite{MI_minimization}.
Whereas, when one wants to maximize the dependence between $\mathbf{u}$ and $\mathbf{v}$, that is, to increase $I(\mathbf{u}, \mathbf{v})$, an approach to maximize the lower bound $I_{\text{InfoNCE}}(\mathbf{u}, \mathbf{v})$ of MI~\cite{MI_maximization} can be used.
% In this work, we use $I_{\text{InfoNCE}}$ and $I_{\text{CLUB}}$ in the design of the loss function for the semantic feature extractor.

%----------------------------------------------------------
\section{SE-GZSL}

% 너무 중복심하면 이건 빼자
% As mentioned, key idea of the proposed SE-GZSL technique is to use the semantic feature containing only attribute-related information for the GZSL classification.
In this section, we present the proposed GZSL technique called semantic feature extraction-based GZSL (SE-GZSL).
We first discuss how to extract the semantic feature from the image feature and then delve into the GZSL classification using the extracted semantic feature.

\subsection{Semantic Feature Extraction}

In extracting the semantic feature from the image feature, the proposed SE-GZSL technique uses the modified autoencoder architecture where two encoders, called semantic and residual encoders, are used in capturing the attribute-related information and the attribute-irrelevant information, respectively (see Fig~\ref{fig:SD}).
As mentioned, in the autoencoder training, we use two loss functions: 1) MI-based loss to encourage the semantic encoder to capture all attribute-related information and 2) similarity-based loss to encourage the semantic encoder not to capture attribute-irrelevant information.
In this subsection, we discuss the overall training loss with emphasis on these two.
% Conventionally, the reconstruction loss is used in the autoencoder training to preserve the input information in the encoder outputs.
% In our approach, to enforce the semantic encoder to capture the attribute-related information exclusively, we additionally use two loss functions: 1) MI-based loss to encourage the semantic encoder to capture all attribute-related information and 2) similarity-based loss to discourage the semantic encoder to capture any unwanted attribute-irrelevant information.

\paragraph{MI-based Loss}

To make sure that all the attribute-related information is contained in the semantic encoder output, we use MI in the autoencoder training.
% Specifically, to encourage the semantic encoder to capture the attribute-related information, we maximize MI between the semantic encoder output and the attribute which is given by manual annotation.
To do so, we maximize MI between the semantic encoder output and the attribute which is given by manual annotation.
At the same time, to avoid capturing of attribute-related information in the residual encoder, we minimize MI between the residual encoder output and the attribute.
Let $\mathbf{z}_{s}$ and $\mathbf{z}_{r}$ be the semantic and residual encoder outputs corresponding to the image feature $\mathbf{x}$, and $\mathbf{a}$ be the image attribute (see Fig.~\ref{fig:SD}). Then, our training objective can be expressed as
\begin{align}
    \text{minimize}~~~-\lambda_{s} I(\mathbf{z}_{s}, \mathbf{a}) + \lambda_{r} I(\mathbf{z}_{r}, \mathbf{a}),
    \label{eq:separation loss_MI form}
\end{align}
where $\lambda_{s}$ and $\lambda_{r}$ ($\lambda_{s}, \lambda_{r} > 0$) are weighting coefficients.

Since the computation of MI is not tractable, we use InfoNCE and CLUB (see~\eqref{eq:MI_lower bound_InfoNCE} and~\eqref{eq:MI_upper bound_CLUB}) as a surrogate of MI.
In our approach, to minimize the objective function in~\eqref{eq:separation loss_MI form}, we express its upper bound using InfoNCE and CLUB and then train the autoencoder in a way to minimize the upper bound.
Using the relationship between MI and its estimators in~\eqref{eq:MI inequality}, the upper bound $\mathcal{L}_{\text{MI}}$ of the objective function in~\eqref{eq:separation loss_MI form} is
\begin{align}
    \mathcal{L}_{\text{MI}}
    &= -\lambda_{s} I_{\text{InfoNCE}}(\mathbf{z}_{s}, \mathbf{a}) + \lambda_{r} I_{\text{CLUB}} (\mathbf{z}_{r}, \mathbf{a}) \nonumber \\
    &= -\lambda_{s} \mathbb{E}_{p(\mathbf{z}_{s}, \mathbf{a})} [ f(\mathbf{z}_{s}, \mathbf{a}) ] \nonumber \\
    &~~~ + \hspace{-.5mm} \lambda_{s} \mathbb{E}_{p(\mathbf{z}_{s})} \hspace{-1mm} \left [ \log \left (
    \mathbb{E}_{p(\mathbf{a})} [ \exp ( f(\mathbf{z}_{s}, \mathbf{a}) ) ] \right ) \right ] \nonumber \\
    &~~~ +\lambda_{r} \left ( 
    \mathbb{E}_{p(\mathbf{z}_{r}, \mathbf{a})} \hspace{-1mm} \left [ 
    \log p(\mathbf{a} | \mathbf{z}_{r} ) \right ] \hspace{-.5mm} - 
    \hspace{-.5mm} \mathbb{E}_{p(\mathbf{z}_{r})p(\mathbf{a})} \hspace{-1mm} \left [ \log p(\mathbf{a} | \mathbf{z}_{r} ) \right ]
    \right ) \hspace{-1mm}. 
    \label{eq:separation loss_expectation form}
\end{align}
Let $\mathcal{Y}_{s}$ be the set of seen classes, $\mathbf{a}_{c}$ be the attribute of a seen class $c \in \mathcal{Y}_{s}$, and $\{ \mathbf{x}_{c}^{(i)} \}_{i=1}^{N_{c}}$ be the set of training image features for the class $c$.
Further, let $\mathbf{z}_{c, s}^{(i)}$ and $\mathbf{z}_{c, r}^{(i)}$ be the semantic and residual encoder outputs corresponding to the input image feature $\mathbf{x}_{c}^{(i)}$, respectively, then $\mathcal{L}_{\text{MI}}$ can be expressed as
\begin{align}
    \mathcal{L}_{\text{MI}}
    &= -\frac{\lambda_{s}}{N} \sum_{c \in \mathcal{Y}_{s}} \sum_{i=1}^{N_{c}} \log 
    \frac{\exp ( f ( \mathbf{z}_{c, s}^{(i)}, \mathbf{a}_{c} ) )}
    {\frac{1}{|\mathcal{Y}_{s}|} \hspace{-.5mm} \underset{c^{\prime} \in \mathcal{Y}_{s}}{\sum} \exp ( f ( \mathbf{z}_{c, s}^{(i)}, \mathbf{a}_{c^{\prime}} ) )} \nonumber \\
    &~+\frac{\lambda_{r}}{N} \hspace{-.5mm} \sum_{c \in \mathcal{Y}_{s}} \hspace{-.5mm} \sum_{i=1}^{N_{c}} \hspace{-.5mm} \left ( \log p(\mathbf{a}_{c} | \mathbf{z}_{c, r}^{(i)}) 
    - \hspace{-2mm} \underset{c^{\prime} \in \mathcal{Y}_{s}}{\sum} \hspace{-.5mm} \frac{\log p(\mathbf{a}_{c^{\prime}} | \mathbf{z}_{c, r}^{(i)})}{|\mathcal{Y}_{s}|}
    \hspace{-.5mm} \right ) \hspace{-1mm},
    \label{eq:separation loss}
\end{align}
where $N = \sum_{c \in \mathcal{Y}_{s}} N_{c}$ is the total number of training image features.

\paragraph{Similarity-based Loss}

% Thus far, we have presented the MI-based loss to encourage the semantic encoder to capture all the attribute-related information.
We now discuss the similarity-based loss to enforce the semantic encoder not to capture any attribute-irrelevant information.

Since images belonging to the same class have the same attribute, attribute-related image features of the same class would be more or less similar. 
This means that if the semantic encoder captures attribute-related information only, then the similarity between semantic encoder outputs of the same class should be large.
% In other words, the similarity between semantic encoder output samples $\mathbf{z}_{c, s}^{(i)}$ of the class $c$ would be large.
Inspired by this observation, to remove the attribute-irrelevant information from the semantic encoder output, we train the semantic encoder in a way to maximize the similarity between outputs of the same class:
\begin{align}
    \text{maximize}~~~\sum_{j=1}^{N_{c}} \exp (\similarity ( \mathbf{z}_{c, s}^{(i)}, \mathbf{z}_{c, s}^{(j)} ) ),
    \label{eq:similarity loss_same class}
\end{align}
where the similarity is measured using the cosine-similarity function defined as
\begin{align*}
    \similarity(\mathbf{u}, \mathbf{v}) 
    &= \frac{\langle \mathbf{u}, \mathbf{v} \rangle}
    {\| \mathbf{u} \|_{2} \| \mathbf{v} \|_{2}}.
\end{align*}
% In doing so, the attribute-irrelevant information contained in a few image samples can be removed from the semantic encoder output.
% Inspired by this, we train the autoencoder to strengthen the similarity between semantic encoder outputs of images for the same class.
% In doing so, some attribute-irrelevant information, contained only in a few image samples, can be removed from the semantic encoder output.
% For a semantic encoder output sample $\mathbf{z}_{c, s}^{(i)}$, our training goal can be expressed as
% We note that in our approach, features containing only attribute-related information (semantic latent vectors) are used for the classification.

Also, we minimize the similarity between semantic encoder outputs of different classes to obtain sufficiently distinct semantic encoder outputs for different classes:
\begin{align}
    \text{minimize}~~~\sum_{c^{\prime} \neq c} \sum_{j=1}^{N_{c^{\prime}}} \exp ( \similarity ( \mathbf{z}_{c, s}^{(i)}, \mathbf{z}_{c^{\prime}, s}^{(j)} ) ).
    \label{eq:similarity loss_different classes}
\end{align}
Using the fact that one can maximize $A$ and minimize $B$ at the same time by minimizing $-\log \frac{1}{1+B/A} = -\log \frac{A}{A+B}$, we obtain the similarity-based loss as
% \footnote{We use the fact that one can maximize $A$ and minimize $B$ by minimizing $-\log \frac{1}{1+B/A} = -\log \frac{A}{A+B}$.}:
\begin{align}
    \mathcal{L}_{\text{sim}}
    &= -\frac{1}{N} \sum_{c \in \mathcal{Y}_{s}} \sum_{i=1}^{N_{c}} \log
    \frac{\underset{j=1}{\overset{N_{c}}{\sum}} \exp \hspace{-.7mm} \left ( \hspace{-.5mm} \similarity ( \mathbf{z}_{c, s}^{(i)}, \mathbf{z}_{c, s}^{(j)} ) \hspace{-.5mm} \right )}
    { \underset{c^{\prime} \in \mathcal{Y}_{s}}{\sum} \underset{j=1}{\overset{N_{c^{\prime}}}{\sum}} \exp \hspace{-.7mm} \left ( \hspace{-.5mm} \similarity ( \mathbf{z}_{c, s}^{(i)}, \mathbf{z}_{c^{\prime}, s}^{(j)} ) \hspace{-.5mm} \right ) }.
    \label{eq:similarity loss}
\end{align}

\subsubsection{Overall Loss} 

By adding the conventional reconstruction loss $\mathcal{L}_{\text{recon}}$ for the autoencoder, the MI-based loss $\mathcal{L}_{\text{MI}}$, and the similarity-based loss $\mathcal{L}_{\text{sim}}$, we obtain the overall loss function as
\begin{align}
    \mathcal{L}_{\text{total}}
    &= \mathcal{L}_{\text{recon}} + \mathcal{L}_{\text{MI}} + \lambda_{\text{sim}} \mathcal{L}_{\text{sim}},
    \label{eq:decomposition loss}
\end{align}
where $\lambda_{\text{sim}}$ is a weighting coefficient and $\mathcal{L}_{\text{recon}}$ is the reconstruction loss given by
\begin{align}
    \mathcal{L}_{\text{recon}}
    &= \frac{1}{N} \sum_{c \in \mathcal{Y}_{s}} \sum_{i=1}^{N_{c}} \| \mathbf{x}_{c}^{(i)} - \widehat{\mathbf{x}}_{c}^{(i)} \|_{2}.
\end{align}
Here, $\widehat{\mathbf{x}}_{c}^{(i)}$ is the image feature reconstructed using the semantic and residual encoder outputs ($\mathbf{z}_{c, s}^{(i)}$ and $\mathbf{z}_{c, r}^{(i)}$) in the decoder.
When the training is finished, we only use the semantic encoder for the purpose of extracting the semantic feature.

\subsection{GZSL Classification Using Semantic Features}

So far, we have discussed how to extract the semantic feature from the image feature.
We now discuss how to perform the GZSL classification using the semantic feature.

In a nutshell, we synthesize semantic feature samples for unseen classes from their attributes.
Once the synthetic samples are generated, the semantic classifier identifying unseen classes from the semantic feature is trained in a supervised manner.
% \footnote{Semantic feature samples for seen classes can be obtained from training image samples by exploiting the semantic encoder.} 

% The proposed SE-GZSL technique performs the GZSL classification using semantic features extracted by the semantic encoder (see Fig.~\ref{fig:overview}).
% To train the semantic classifier that identifies the image class from semantic features, we generate synthetic semantic feature samples for unseen classes.
% Once the synthetic samples for unseen classes are generated, the semantic classifier can be trained easily in a supervised manner.\footnote{We note that semantic feature samples for seen classes can be obtained by exploiting image features of seen classes as inputs to the semantic encoder.} 

\paragraph{Semantic Feature Generation}

% -----------------------------------------------------------------------------
\begin{figure*}[h] 
    \centering
    \centerline{\includegraphics[width=13cm]{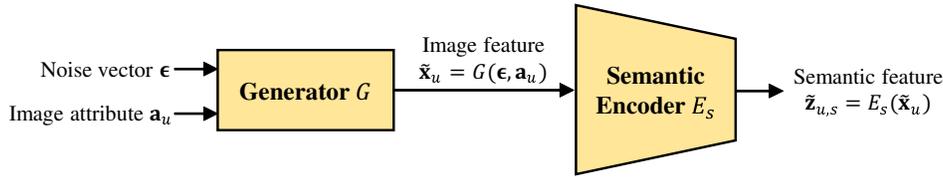}}
    \caption{Illustration of the synthetic semantic feature generation for unseen classes.}
    \label{fig:generator}
\end{figure*}
% ---------------------------------------------------------------------

To synthesize the semantic feature samples for unseen classes, we first generate image features from the attributes of unseen classes and then extract the semantic features from the synthetic image features using the semantic encoder (see Fig.~\ref{fig:generator}).

In synthesizing the image feature, we employ WGAN that mitigates the unstable training issue of GAN by exploiting a Wasserstein distance-based loss function~\cite{WGAN}. 
The main component in WGAN is a generator $G$ synthesizing the image feature $\widetilde{\mathbf{x}}_{c}$ from a random noise vector $\boldsymbol{\epsilon} \sim \mathcal{N}(\mathbf{0}, \mathbf{I})$ and the image attribute $\mathbf{a}_{c}$ (i.e., $\widetilde{\mathbf{x}}_{c} = G(\boldsymbol{\epsilon}, \mathbf{a}_{c})$).
Conventionally, WGAN is trained to minimize the Wasserstein distance between the distributions of real image feature $\mathbf{x}_{c}$ and generated image feature $\widetilde{\mathbf{x}}_{c}$ given by
\begin{align}
    \lefteqn{\hspace{-.2cm} \mathcal{L}_{G, \text{WGAN}}} \nonumber \\
    &\hspace{-.3cm}= \max_{D} \hspace{-.5mm} \bigg ( 
    \mathbb{E}_{p(\mathbf{x}_{c} | \mathbf{a}_{c})} [ D(\mathbf{x}_{c}, \mathbf{a}_{c}) ] 
    - \mathbb{E}_{p(\widetilde{\mathbf{x}}_{c} | \mathbf{a}_{c})} [ D(\widetilde{\mathbf{x}}_{c}, \mathbf{a}_{c}) ]  \nonumber \\
    &\hspace{-.3cm}~~~~~~~~~~~~~~ -\lambda_{\text{gp}} \mathbb{E}_{p(\widehat{\mathbf{x}}_{c} | \mathbf{a}_{c})} \hspace{-.8mm} \left [ \left ( \| \nabla_{\widehat{\mathbf{x}}_{c}} D(\widehat{\mathbf{x}}_{c}, \mathbf{a}_{c}) \|_{2} - 1 \right )^{2} 
    \right ] \hspace{-.8mm} \bigg ),
    \label{eq:generator_WGAN loss}
\end{align}
where $D$ is an auxiliary network (called critic), $\widehat{\mathbf{x}}_{c} = \alpha \mathbf{x}_{c} + (1 - \alpha) \widetilde{\mathbf{x}}_{c}~(\alpha \sim \mathcal{U}(0, 1))$, and $\lambda_{\text{gp}}$ is the regularization coefficient (a.k.a., gradient penalty coefficient)~\cite{WGAN-GP}.
% The first two terms in~\eqref{eq:generator_WGAN loss} represent the Wasserstein distance between the real and generated data distributions~\cite{WGAN}, which implies that real-like image features can be generated by minimizing $\mathcal{L}_{G, \text{WGAN}}$. 
In our scheme, to make sure that the semantic feature $\widetilde{\mathbf{z}}_{c, s}$ obtained from $\widetilde{\mathbf{x}}_{c}$ is similar to the real semantic feature $\mathbf{z}_{c, s}$, we additionally use the following losses in the WGAN training:
\begin{align}
    \mathcal{L}_{G, \text{MI}}
    &= -I_{\text{InfoNCE}}(\widetilde{\mathbf{z}}_{c, s}, \mathbf{a}_{c}), 
    \label{eq:generator_InfoNCE loss} \\
% \end{align}
% \begin{align}
    \mathcal{L}_{G, \text{sim}}
    &= -\mathbb{E}_{p(\widetilde{\mathbf{z}}_{c, s})} \hspace{-1mm} \left [
    \log \frac{\underset{i=1}{\overset{N_{c}}{\sum}} \exp ( \similarity (\widetilde{\mathbf{z}}_{c, s}, \mathbf{z}_{c, s}^{(i)}) )}
    {\underset{c^{\prime}=1}{\overset{S}{\sum}} \underset{i=1}{\overset{N_{c^{\prime}}}{\sum}} \exp ( \similarity (\widetilde{\mathbf{z}}_{c, s}, \mathbf{z}_{c^{\prime}, s}^{(i)}) )}
    \right ].
    \label{eq:generator_similarity loss}
\end{align}
We note that these losses are analogous to the losses with respect to the real semantic feature $\mathbf{z}_{c, s}$ in~\eqref{eq:separation loss_expectation form} and~\eqref{eq:similarity loss}, respectively.
By combining~\eqref{eq:generator_WGAN loss},~\eqref{eq:generator_InfoNCE loss}, and~\eqref{eq:generator_similarity loss}, we obtain the overall loss function as
\begin{align}
    \mathcal{L}_{G}
    &= \mathcal{L}_{G, \text{WGAN}} + \lambda_{G, \text{MI}} \mathcal{L}_{G, \text{MI}} + \lambda_{G, \text{sim}} \mathcal{L}_{G, \text{sim}}, \label{eq:generator loss}
\end{align}
where $\lambda_{G, \text{MI}}$ and $\lambda_{G, \text{sim}}$ are weighting coefficients. 

After the WGAN training, we use the generator $G$ and the semantic encoder $E_{s}$ in synthesizing semantic feature samples of unseen classes.
Specifically, for each unseen class $u \in \mathcal{Y}_{u}$, we generate the semantic feature $\widetilde{\mathbf{z}}_{u, s}$ by synthesizing the image feature $\widetilde{\mathbf{x}}_{u} = G(\boldsymbol{\epsilon}, \mathbf{a}_{u})$ using the generator and then exploiting it as an input to the semantic encoder (see Fig.~\ref{fig:generator}):
\begin{align}
    \widetilde{\mathbf{z}}_{u, s}
    = E_{s}(\widetilde{\mathbf{x}}_{u})
    = E_{s}(G(\boldsymbol{\epsilon}, \mathbf{a}_{u})).
\end{align}
By resampling the noise vector $\boldsymbol{\epsilon} \sim \mathcal{N}(\mathbf{0}, \mathbf{I})$, a sufficient number of synthetic semantic features can be generated.

\paragraph{Semantic Feature-based Classification}

% \begin{figure*}[!t] 
%     \centering
%     \centerline{\includegraphics[width=14cm]{overview.eps}}
%     \caption{Illustration of the image classifier in the proposed SE-GZSL technique.}
%     \label{fig:overview}
% \end{figure*}

After generating synthetic semantic feature samples for all unseen classes, we train the semantic feature classifier using a supervised learning model (e.g., softmax classifier, support vector machine, and nearest neighbor). 
Suppose, for example, that the softmax classifier is used as a classification model.
Let $\{ \widetilde{\mathbf{z}}_{u, s}^{(i)} \}_{i=1}^{N_{u}}$ be the set of synthetic semantic feature samples for the unseen class $u$, then the semantic feature classifier is trained to minimize the cross entropy loss\footnote{We recall that $\{ \mathbf{z}_{c, s}^{(i)} \}_{i=1}^{N_{c}}$ is the set of semantic features for the seen class $c \in \mathcal{Y}_{s}$.}
\begin{align}
    \mathcal{L}_{\text{CE}}
    &= -\sum_{c \in \mathcal{Y}_{s}} \sum_{i=1}^{N_{c}} \log P( c | \mathbf{z}_{c, s}^{(i)} ) - \sum_{u \in \mathcal{Y}_{u}} \sum_{i=1}^{N_{u}} \log P( u | \widetilde{\mathbf{z}}_{u, s}^{(i)} ), 
\end{align}
where 
\begin{align}
    P(y|\mathbf{z}) 
    &= \frac{\exp (\mathbf{w}_{y}^{T} \mathbf{z} + b_{y})}{ \sum_{y^{\prime} \in \mathcal{Y}_{s} \cup \mathcal{Y}_{u}} \exp (\mathbf{w}_{y^{\prime}}^{T} \mathbf{z} + b_{y^{\prime}})}
\end{align}
and $\mathbf{w}_{y}$ and $b_{y}$ are weight and bias parameters of the softmax classifier to be learned.

\subsection{Comparison with Conventional Approaches}

There have been previous efforts to extract the semantic feature from the image feature~\cite{DLFZRL, RFF-GZSL, Disentangled-VAE, SDGZSL}.
While our approach seems to be a bit similar to~\cite{Disentangled-VAE} and~\cite{SDGZSL} in the sense that the autoencoder-based image feature decomposition method is used for the semantic feature extraction, our work is dearly distinct from those works in two respects.
First, we use different training strategy in capturing the attribute-related information.
In our approach, to make sure that the semantic encoder output contains all the attribute-related information, we use two complementary loss terms: 1) the loss term to encourage the semantic encoder to capture the attribute-related information and 2) the loss term to discourage the residual encoder to capture any attribute-related information (see~\eqref{eq:separation loss_MI form}).
Whereas, the training loss used to remove the attribute-related information from the residual encoder output has not been used in~\cite{Disentangled-VAE, SDGZSL}.
Also, we employ a new training loss $\mathcal{L}_{\text{sim}}$ to remove the attribute-irrelevant information from the semantic encoder output (see~\eqref{eq:similarity loss}), for which there is no counterpart in~\cite{Disentangled-VAE, SDGZSL}.

% %-----------------------------------------------
% \begin{table}[t]

% \centering

% {\resizebox{1.0\linewidth}{!}{
% \begin{tabular}{c  c  c  c}
% \toprule
% Dataset &
% $| \mathcal{Y}^{s} |$ (Train + Val.) & 
% $| \mathcal{Y}^{u} |$ &
% \# of Attributes  \\
% \midrule
% AwA1 &  
% 27 + 13 &
% 10 &
% 85 \\
% AwA2 &  
% 27 + 13 &
% 10 &
% 85 \\
% CUB &  
% 100 + 50 &
% 50 &
% 312 \\
% SUN &  
% 580 + 65 &
% 72 &
% 102 \\
% \bottomrule
% \end{tabular}
% }}

% \caption{Statistics of the datasets used in our experiments.}

% \label{tab:datasets}
% \end{table}
% %-----------------------------------------------

%-------------------------------------------------------
\section{Experiments}

%-----------------------------------------------
\begin{table}[t]

\centering

{\resizebox{1.0\linewidth}{!}{
\begin{tabular}{c | c | c | c | c }
\toprule
Classifier input    &
AwA1    &
AwA2    & 
CUB     &
SUN     \\
\midrule
Image feature &  
90.9    &
92.8    &
73.8    &
47.1    \\
Semantic feature &
\textbf{91.9}    &
\textbf{93.4}    &
\textbf{76.1}    &
\textbf{49.3}    \\
\bottomrule
\end{tabular}
}}

\caption{Top-1 accuracy of image feature-based and semantic feature-based image classifiers.}

\label{tab:result_effect of attribute-related feature extraction}
\end{table}

\begin{figure*}[!t]
\begin{minipage}[b]{0.33 \linewidth}
  \centering
  \centerline{\includegraphics[width=6cm]{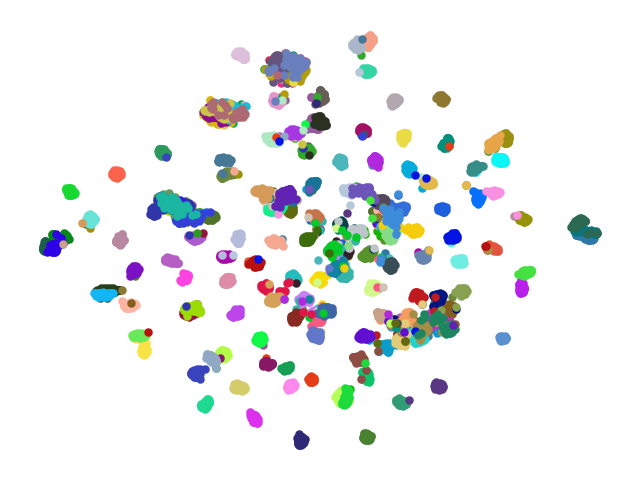}}
%  \vspace{1.5cm}
  \centerline{(a) Semantic features}
\end{minipage}
\hfill
\begin{minipage}[b]{0.33 \linewidth}
  \centering
  \centerline{\includegraphics[width=6cm]{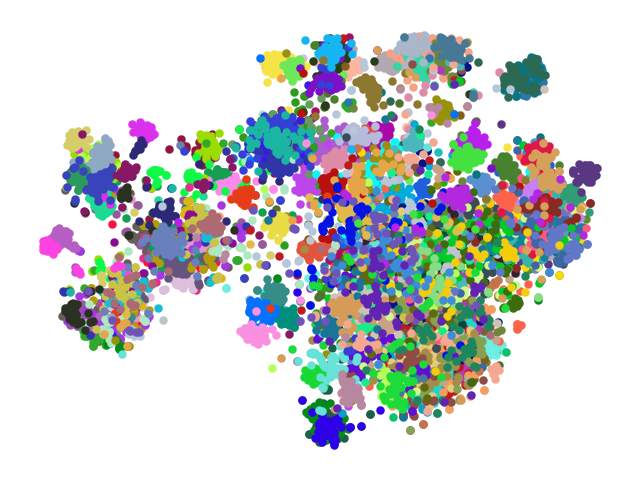}}
%  \vspace{1.5cm}
  \centerline{(b) Image features}
\end{minipage}
\hfill
\begin{minipage}[b]{0.33 \linewidth}
  \centering
  \centerline{\includegraphics[width=6cm]{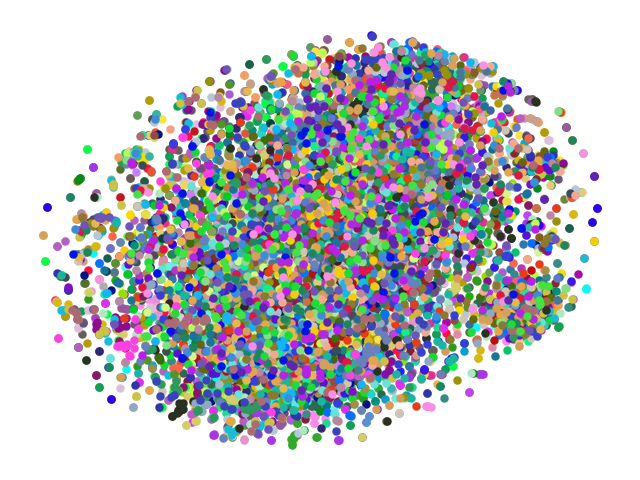}}
%  \vspace{1.5cm}
  \centerline{(c) Residual features}
\end{minipage}

\caption{t-SNE visualization of (a) semantic features, (b) image features, and (c) residual features. Samples for the same class are indicated in the same color.}

\label{fig:t-SNE}
\end{figure*}

\begin{table*}[t]

\centering

{\resizebox{1.0\linewidth}{!}{
\begin{tabular}{c | c | c c c | c c c | c c c | c c c}
\toprule
\multirow{2}{2.2cm}{\centering Method} &
\multirow{2}{2.2cm}{\centering Feature Type} &
\multicolumn{3}{c|}{AwA1} & 
\multicolumn{3}{c|}{AwA2} & 
\multicolumn{3}{c|}{CUB} & 
\multicolumn{3}{c}{SUN} \\
\cline{3-14}
&
&
$acc_{s}$ & $acc_{u}$ & $acc_{h}$ &
$acc_{s}$ & $acc_{u}$ & $acc_{h}$ &
$acc_{s}$ & $acc_{u}$ & $acc_{h}$ &
$acc_{s}$ & $acc_{u}$ & $acc_{h}$ \\
\toprule
% ALE &  
% \multirow{2}{2.2cm}{\centering ResNet} &
% 76.1  &  16.8  &  27.5  &
% 81.8  &  14.0  &  23.9  &
% 62.8  &  23.7  &  34.4  &
% 33.1  &  21.8  &  26.3 \\
% DeViSE &  
% &
% 68.7  &  13.4  &  22.4  &
% 74.7  &  17.1  &  27.8  &
% 53.0  &  23.8  &  32.8  &
% 27.4  &  16.9  &  20.9 \\
% \midrule
CVAE-GZSL &  
\multirow{9}{2.2cm}{\centering ResNet} &
-  &  -  &  47.2  &
-  &  -  &  51.2  &
-  &  -  &  34.5  &
-  &  -  &  26.7 \\
% SE-GZSL &  
% 67.8  &  56.3  &  61.5  &
% 68.1  &  58.3  &  62.8  &
% 53.3  &  41.5  &  46.7  &
% 30.5  &  40.9  &  34.9 \\
f-CLSWGAN &
&
61.4  &  57.9  &  59.6 &
-  &  -  &  -  &
57.7  &  43.7  &  49.7 &
36.6  &  42.6  &  39.4 \\
cycle-CLSWGAN &  
&
64.0  &  56.9  &  60.2  &
-  &  -  &  -  &
61.0  &  45.7  &  52.3  &
33.6  &  49.4  &  40.0 \\
f-VAEGAN-D2 &
&
-  &  -  &  -  &
70.6  &  57.6  &  63.5  &
60.1  &  48.4  &  53.6  &
38.0  &  45.1  &  41.3 \\
LisGAN &
&
76.3  &  52.6  &  62.3 &
-  &  -  &  -  &
57.9  &  46.5  &  51.6 &
37.8  &  42.9  &  40.2 \\
CADA-VAE &
&
72.8  &  57.3  &  64.1 &
75.0  &  55.8  &  63.9 &
53.5  &  51.6  &  52.4 &
35.7  &  47.2  &  40.6 \\
DASCN &
&
68.0  &  59.3  &  63.4 &
-  &  -  &  -  &
59.0  &  45.9  &  51.6 &
38.5  &  42.4  &  40.3 \\
LsrGAN &
&
74.6  &  54.6  &  63.0  &
-  &  -  &  -  &
59.1  &  48.1  &  53.0  &
37.7  &  44.8  &  40.9 \\
Zero-VAE-GAN &
&
66.8  &  58.2  &  62.3  &
70.9  &  57.1  &  62.5  &
47.9  &  43.6  &  45.5  &
30.2  &  45.2  &  36.3 \\
\midrule
DLFZRL &
\multirow{3}{2.2cm}{\centering Semantic} &
-  &  -  &  61.2 &
-  &  -  &  60.9 &
-  &  -  &  51.9 &
-  &  -  &  {\underline{42.5}} \\
RFF-GZSL &
&
75.1  &  59.8  &  {\underline{66.5}} &
-  &  -  &  - &
56.6  &  52.6  &  {\underline{54.6}} &
38.6  &  45.7  &  41.9 \\
Disentangled-VAE &
&
72.9  &  60.7  &  66.2 &
80.2  &  56.9  &  {\underline{66.6}} &
58.2  &  51.1  &  54.4 &
36.6  &  47.6  &  41.4 \\
\midrule
{\bf{SE-GZSL}} &
Semantic &
76.7  &  61.3  & {\bf{68.1}} &
80.7  &  59.9  & {\bf{68.8}} &
60.3  &  53.1  & {\bf{56.4}} &
40.7  &  45.8  & {\bf{43.1}}  \\
\bottomrule
\end{tabular}
}}

\caption{GZSL classification performance of the proposed SE-GZSL technique and conventional approaches. `-' means that the result is not reported in the references. The best results are in bold, and the second best results are underlined.}

\label{tab:result_GZSL performance}
\end{table*}
%-----------------------------------------------

\begin{table*}[t]

\centering

{\resizebox{1.0\linewidth}{!}{
\begin{tabular}{c | c c c | c c c | c c c | c c c}
\toprule
\multirow{2}{2.2cm}{\centering Loss}& 
\multicolumn{3}{c|}{AwA1} & 
\multicolumn{3}{c|}{AwA2} & 
\multicolumn{3}{c|}{CUB} & 
\multicolumn{3}{c}{SUN} \\
\cline{2-13}
&
$acc_{s}$ & $acc_{u}$ & $acc_{h}$ &
$acc_{s}$ & $acc_{u}$ & $acc_{h}$ &
$acc_{s}$ & $acc_{u}$ & $acc_{h}$ &
$acc_{s}$ & $acc_{u}$ & $acc_{h}$ \\
\toprule
$\mathcal{L}_{\text{recon}}$ &
64.6    &   53.1    &   58.3    &
68.9    &   55.7    &   61.6    &
54.5    &   46.1    &   49.9    &
38.4    &   40.6    &   39.4    \\
$\mathcal{L}_{\text{recon}}$ + $\mathcal{L}_{\text{MI}}$ &
75.0    &   57.9    &   65.4    &   
74.2    &   58.6    &   65.5    &
59.4    &   51.5    &   55.1    &
37.1    &   46.5    &   41.3    \\
$\mathcal{L}_{\text{recon}}$ + $\mathcal{L}_{\text{MI}}$ + $\mathcal{L}_{\text{sim}}$ &
76.7    &   61.3    &   \textbf{68.1}    &
80.7    &   59.9    &   \textbf{68.8}    &
60.3    &   53.1    &   \textbf{56.4}    &
40.7    &   45.8    &   \textbf{43.1}    \\
\bottomrule
\end{tabular}
}}

\caption{Ablation study on the performance of SE-GZSL.}

\label{tab:result_ablation study}
\end{table*}
%-----------------------------------------------

\subsection{Experimental Setup}

\paragraph{Datasets}

In our experiments, we evaluate the performance of our model using four benchmark datasets: AwA1, AwA2, CUB, and SUN. 
The AwA1 and AwA2 datasets contain 50 classes of animal images annotated with 85 attributes~\cite{zsl_proposal, AwA2}.
% Each class is annotated with 85 attributes, and training images are available only for 40 classes.
The CUB dataset contains 200 species of bird images annotated with 312 attributes~\cite{CUB}.
% among which only 150 species have training image samples. 
% Each bird species is annotated with 312 attributes. 
The SUN dataset contains 717 classes of scene images annotated with 102 attributes~\cite{SUN}. 
% In the training phase, image samples are given only for 645 classes. 
In dividing the total classes into seen and unseen classes, we adopt the conventional dataset split presented in~\cite{AwA2}.
% In Table~\ref{tab:datasets}, we summarize the statistics of each dataset.

\paragraph{Implementation Details} 
As in~\cite{CLSWGAN, CADA-VAE}, we use ResNet-101~\cite{ResNet} as a pre-trained classification network and fix it in our training process.
We implement all the networks in SE-GZSL (semantic encoder, residual encoder, and decoder in the image feature decomposition network, and generator and critic in WGAN) using the multilayer perceptron (MLP) with one hidden layer as in~\cite{CLSWGAN, f-vaegan-d2}.
We set the number of hidden units to 4096 and use LeakyReLU with a negative slope of 0.02 as a nonlinear activation function.
For the output layer of the generator, the ReLU activation is used since the image feature extracted by ResNet is non-negative.
As in~\cite{InfoNCE}, we define the score function $f$ in~\eqref{eq:separation loss_expectation form} as $f(\mathbf{z}_{s}, \mathbf{a}) = \mathbf{z}_{s}^{T} \mathbf{W} \mathbf{a}$ where $\mathbf{W}$ is a weight matrix to be learned.
Also, as in~\cite{CLUB}, we approximate the conditional PDF $p(\mathbf{a} | \mathbf{z}_{r})$ in~\eqref{eq:separation loss_expectation form} using a variational encoder consisting of two hidden layers.
The gradient penalty coefficient in the WGAN loss $\mathcal{L}_{G, \text{WGAN}}$ is set to $\lambda_{\text{gp}} = 10$ as suggested in the original WGAN paper~\cite{WGAN-GP}.
We set the weighting coefficients in~\eqref{eq:separation loss}, ~\eqref{eq:decomposition loss}, and~\eqref{eq:generator loss} to $\lambda_{s}=20, \lambda_{r}=50, \lambda_{\text{sim}}=1, \lambda_{G, \text{MI}}=1, \lambda_{G, \text{sim}} = 0.025$.

\subsection{Semantic Feature-based Image Classification}

We first investigate whether the image classification performance can be improved by exploiting the semantic feature.
To this end, we train two image classifiers: the classifier exploiting the image feature and the classifier utilizing the semantic feature extracted by the semantic encoder. 
To compare the semantic feature directly with the image feature, we use the simple softmax classifier as a classification model.
In Table~\ref{tab:result_effect of attribute-related feature extraction}, we summarize the top-1 classification accuracy of each classifier on test image samples for seen classes.
We observe that the semantic feature-based classifier outperforms the image feature-based classifier for all datasets.
In particular, for the SUN and CUB datasets, the semantic feature-based classifier achieves about $2\%$ improvement in the top-1 classification accuracy over the image feature-based classifier, which demonstrates that the image classification performance can be enhanced by removing the attribute-irrelevant information in the image feature.

\subsection{Visualization of Semantic Features}

In Fig.~\ref{fig:t-SNE}, we visualize semantic feature samples obtained from the CUB dataset using a t-distributed stochastic neighbor embedding (t-SNE), a tool to visualize high-dimensional data in a two-dimensional plane~\cite{t-SNE}.
For comparison, we also visualize image feature samples and residual feature samples extracted by the residual encoder.
We observe that semantic feature samples containing only attribute-related information are well-clustered, that is, samples of the same class are grouped and samples of different classes are separated (see Fig.~\ref{fig:t-SNE}(a)).
Whereas, image feature samples of different classes are not separated sufficiently (see Fig.~\ref{fig:t-SNE}(b)) and residual feature samples are scattered randomly (see Fig.~\ref{fig:t-SNE}(c)).

\subsection{Comparison with State-of-the-art}

We next evaluate the GZSL classification performance of the proposed approach using the standard evaluation protocol presented in~\cite{AwA2}.
Specifically, we measure the average top-1 classification accuracies $acc_{s}$ and $acc_{u}$ on seen and unseen classes, respectively, and then use their harmonic mean $acc_{h}$ as a metric to evaluate the performance.
% \begin{align*}
%     acc_{h} 
%     &= \frac{2 \cdot acc_{s} \cdot acc_{u}}{acc_{s} + acc_{u}}.
% \end{align*}
In Table~\ref{tab:result_GZSL performance}, we summarize the performance of SE-GZSL on different datasets.
For comparison, we also summarize the performance of conventional methods among which DLFZRL, RFF-GZSL, and Disentangled-VAE are semantic feature-based approaches~\cite{DLFZRL, RFF-GZSL, Disentangled-VAE} and other methods are image feature-based approaches~\cite{CVAE-GZSL, CLSWGAN, cycle-WGAN, f-vaegan-d2, LisGAN, CADA-VAE, DASCN, LsrGAN, Zero-VAE-GAN}.

% From the results, we observe that generative model-based approaches outperform compatibility-based approaches by a large margin.
% We also observe that the proposed SE-GZSL outperforms the conventional GZSL approaches exploiting the image feature by a large margin.
From the results, we observe that the proposed SE-GZSL outperforms conventional image feature-based approaches by a large margin.
For example, for the AwA2 dataset, SE-GZSL achieves about 5\% improvement in the harmonic mean accuracy over image feature-based approaches.
We also observe that SE-GZSL outperforms existing semantic feature-based approaches for all datasets.
For example, for the AwA1, AwA2, and CUB datasets, our model achieves about 2\% improvement in the harmonic mean accuracy over the state-of-the-art approaches.

\subsection{Ablation Study}

\paragraph{Effectiveness of Loss Functions}
In training the semantic feature extractor, we have used the MI-based loss $\mathcal{L}_{\text{MI}}$ and the similarity-based loss $\mathcal{L}_{\text{sim}}$.
To examine the impact of each loss function, we measure the performance of three different versions of SE-GZSL: 1) SE-GZSL trained only with the reconstruction loss $\mathcal{L}_{\text{recon}}$, 2) SE-GZSL trained with $\mathcal{L}_{\text{recon}}$ and $\mathcal{L}_{\text{MI}}$, and 3) SE-GZSL trained with $\mathcal{L}_{\text{recon}}$, $\mathcal{L}_{\text{MI}}$, and $\mathcal{L}_{\text{sim}}$.
From the results in Table~\ref{tab:result_ablation study}, we observe that the performance of SE-GZSL can be enhanced greatly by exploiting the MI-based loss $\mathcal{L}_{\text{MI}}$. 
In particular, for the AwA1 and CUB datasets, we achieve more than $5\%$ improvement in the harmonic mean accuracy by utilizing $\mathcal{L}_{\text{MI}}$. 
Also, for the AwA2 dataset, we achieve about $4\%$ improvement of the accuracy.
One might notice that when $\mathcal{L}_{\text{MI}}$ is not used, SE-GZSL performs worse than conventional image feature-based methods (see Table~\ref{tab:result_GZSL performance}).
This is because the semantic encoder cannot capture all the attribute-related information without $\mathcal{L}_{\text{MI}}$, and thus using the semantic encoder output in the classification incurs the loss of the attribute-related information.
We also observe that the performance of SE-GZSL can be improved further by exploiting the similarity-based loss $\mathcal{L}_{\text{sim}}$.
For example, for the AwA2 dataset, more than $3\%$ improvement in the harmonic mean accuracy can be achieved by utilizing $\mathcal{L}_{\text{sim}}$.

\paragraph{Importance of Residual Encoder}

\begin{table}[t]

\centering

{\resizebox{1.0\linewidth}{!}{
\begin{tabular}{c | c | c | c | c}
\toprule
Method &  
AwA1 & 
AwA2 & 
CUB & 
SUN \\
\toprule
SE-GZSL w/o residual encoder &
66.7  &
67.5  &
55.1  &
42.1  \\
\midrule
SE-GZSL w/ residual encoder &
{\bf{68.1}} &
{\bf{68.8}} &
{\bf{56.4}} &
{\bf{43.1}}  \\
\bottomrule
\end{tabular}
}}

\caption{Harmonic mean accuracy of SE-GZSL with and without the residual encoder.}

\label{tab:rebuttal_residual encoder}
\end{table}
%-----------------------------------------------

For the semantic feature extraction, we have decomposed the image feature into the attribute-related feature and the attribute-irrelevant feature using the semantic and residual encoders.
An astute reader might ask why the residual encoder is needed to extract the semantic feature.
To answer this question, we measure the performance of SE-GZSL without using the residual encoder.
From the results in Table~\ref{tab:rebuttal_residual encoder}, we can observe that the GZSL performance of SE-GZSL is degraded when the residual encoder is not used.
This is because if the residual encoder is removed, then the attribute-irrelevant information, required for the reconstruction of the image feature, would be contained in the semantic encoder output and therefore mess up the process to learn the relationship between the image feature and the attribute.

\section{Conclusion}

In this paper, we presented a new GZSL technique called SE-GZSL.
Key idea of the proposed SE-GZSL is to exploit the semantic feature in learning the relationship between the image and the attribute, removing the interference caused by the attribute-irrelevant information.
To extract the semantic feature, we presented the autoencoder-based image feature decomposition network consisting of semantic and residual encoders.
In a nutshell, the semantic and residual encoders capture the attribute-related information and the attribute-irrelevant information, respectively.
In training the image feature decomposition network, we used MI-based loss to encourage the semantic encoder to capture all the attribute-related information and similarity-based loss to discourage the semantic encoder to capture any attribute-irrelevant information.
Our experiments on various datasets demonstrated that the proposed SE-GZSL outperforms conventional GZSL approaches by a large margin.

\section{Acknowledgements}

This work was supported in part by the Samsung Research Funding \& Incubation Center for Future Technology of Samsung Electronics under Grant SRFC-IT1901-17 and in part by the National Research Foundation of Korea (NRF) grant funded by the
Korea government (MSIT) under Grant 2020R1A2C2102198.

{\small
\bibliography{references}
}

\end{document}